# Food Recipe Recommendation Based on Ingredients Detection Using Deep Learning

Md. Shafaat Jamil Rokon, Md Kishor Morol, Ishra Binte Hasan, A. M. Saif, Rafid Hussain Khan
Department of Computer Science
American International University-Bangladesh (AIUB)
{shafaatjamilrokon, kishoremorol, ishrahasan12, alimdsaif1996, rafidkhan321.rk}@gmail.com

## ABSTRACT

Food is essential for human survival, and people always try to taste different types of delicious recipes. Frequently, people choose food ingredients without even knowing their names or pick up some food ingredients that aren't obvious to them from a grocery store. Knowing which ingredients can be mixed to make a delicious food recipe is essential. Selecting the right recipe by choosing a list of ingredients is very difficult for a beginner cook. However, it can be a problem even for experts. There is the constant use of machine learning in our everyday lives. One such example is recognizing objects through image processing. Although this process is complex due to different food ingredients, traditional approaches will lead to an inaccuracy rate. These problems can be solved by machine learning and deep learning approaches. In this paper, we implemented a model for food ingredients recognition and designed an algorithm for recommending recipes based on recognized ingredients. We made a custom dataset consisting of 9856 images belonging to 32 different food ingredients classes. Convolution Neural Network (CNN) model was used to identify food ingredients, and for recipe recommendations, we have used machine learning. We achieved an accuracy of 94%, which is quite impressive.

## 1 INTRODUCTION

People nowadays become very much health conscious and, they try to take at most healthy food in their meal. An appropriate food recipe must require serving the real taste of the food and a balanced diet, and healthy life. As we all know, health is wealth, so food selection is essential for better health. Sometimes people pick some ingredient that they do not even know their names. Or pick up some ingredients from a grocery store, but they do not know how to make recipes using those ingredients. People should know which ingredients can be mixed and make delicious recipes. For a beginner, it is very challenging to select a recipe just by seeing the ingredients. Even for an expert cook, it is also problematic. In this paper, we try to solve this problem. We implemented an approach of recognizing food ingredients, and after analyzing and classifying these recognized ingredients, food that can be cooked with compatible recipes will be recommended. This recommendation will help people to get a suitable food recipe. Online cooking recipe sites have become very popular, but they sometimes get confused when people want to cook by following these sites. This may cause to stop users from referring to these types of cooking sites throughout searching at grocery stores. Technology has to be improved to solve these problems, and then object recognition technology was introduced to the world. There has been significant progress in object recognition technology. Object recognition is a technology that identifies the objects shown in an image based on their specific properties [1]. So, Object recognition is a process that finds things in the real world from analyzing an image. To use object recognition, computer vision, and deep learning models, open-source software libraries such as the Open Computer Vision Library (OpenCV) [2], TensorFlow [3], NumPy [4], and Keras [5] have been used widely. With these libraries, we can implement object recognition schemes on PCs and mobile devices such as iPhones and Android smartphones. With the recent progress of mobile devices and their explosive spread, we have now been able to recognize objects at any time on a mobile device. OpenCV [2] can be used in image processing and representation. Deep learning APIs like NumPy can do the numerical computation, making machine learning faster and easier. TensorFlow[3] API can be used to recognize the image and train deep learning architectures for image classification [4]. There are pre-trained deep learning architectures such as MobileNet [6], ResNet [7], and Inception [8], VGG16 [9] can be used too. The advantage of using pre-trained architecture is that we can modify it using transfer learning [10]. With the help of these libraries, APIs, and architectures, we can easily recognize objects from an image.

For recipes identification, very little work has been done because of not having a proper dataset, and also, working on multiple object recognizing is still challenging. Most of the papers we found are just on recipes identification or food recommendation. But we aim to detect food ingredients by reaching out maximum classification rate and recommending worthy food recipes. In the following section we discuss about related previous works.

## 2 RELATED WORKS

Food detection, ingredient detection, or recipe recommendation have received increasing attention in recent years. All these profound deep learning-based works related to object detection and classification. There are adequate deep learning models and methods available for object detection and classification. Use of color histogram, BoF, linear kernel SVM classifier, and K-nearest neighbors are well known.

Kawano et al. [12] presented an approach for a real-time food recognition system for smartphones. Two types of real-time image recognition methods have been used. One is the combination of bag-of-features (BoF) and color histograms with X2 kernel feature maps. And the second one is the HOG patch descriptor and a color patch descriptor with the state- of-the-art Fisher Vector representation. Linear SVM has been used as a classifier. They have achieved an accuracy of 79.2% classification rate.

Bolaños et al. [13], used CNN for food image recognition. They use two different inputs for their method. The first one provides a low-level description of the food image. The penultimate layer of the InceptionResNetV2 CNN is used in the first method, and the second one provides a high-level description of the food image using LogMeals API. Three different CNNs were supplied by this LogMeals API that predicts food groups, ingredients, and dishes.

Chang Liu et al. [14], proposed CNN based novel approach for visual-based food image recognition with 7-layer architecture and achieved 93.7% top-5 accuracy using the UEC-100 dataset were existing approaches like SURF- BoF + Color Histogram and MKL could achieve 68.3% to 76.8% top-5 accuracy with the same dataset.

Raboy-McGowan and L. Gonzalez, et al. [15][16] both used the Recipe 1M dataset and introduced Recipe Net, a food to recipe generator trained on the Recipe 1M dataset. They have used ResNet-50, DenseNet-121 convolutional neural networks tclassify food images and encode their features. After that, they used K-nearest neighbors to recommend recipes from the Recipe 1M dataset.

Chen, J. et al. [17] used a dataset composed of 61,139 image-recipe pairs gathered from the Go Cooking website. They have explored the recent advances in cross-modality learning to address the problems as mentioned earlier. A deep model stacked attention network (SAN) is absorbed and revised in their system.

KeijiYanai, et al. [18] made a food ingredient recognition system for android OS. The system used a color-histogram-based bag-of-features pull out from several frames as an image file and a linear kernel SVM as a classifier. They stored 30 videos in the database, and their accuracy was 83.93%.

Suyash Maheshwari et al. [19] proposed two algorithms to recommend ingredients. Cooking and tasting experiments showed the proposed methods were effective for each purpose. Through cooking demonstration with the recommended alternative ingredients and subjective evaluation experiments, it was confirmed that both algorithms recommended acceptable ingredients for over 90 percent. However, algorithm one might not recommend an alternative ingredient similar to the exchange ingredient.

Mona Mishra, Yifan Gong et al. [20] for their work, thousands of images are trained for various categories to build a CNN for image recognition. And predict what the object in the image from users upload.

So, we understood that we have a good scope of making a model that can identify ingredients and then recommend recipes based on those ingredients. In the following section, we will discuss our proposed model.

## 3 METHODOLOGY

### 3.1 Proposed Method

First, we divided our work into two parts. One is identifying 32 ingredients where we used Convolution Neural Network (CNN), and after getting the result, we proceed to the second part. Secord part is to the recipe recommendations containing 19 different classes.

We have used Transfer Learning for training our CNN model for the first part. Reuse of a pre-trained model is known as transfer learning. As a pre-trained model, we have used ResNet50. Transfer learning is immensely popular in deep learning because it can train deep neural networks with relatively little data [11]. Transfer learning is efficient since most real-world problems don't have millions of labeled data points to train such complex deep neural networks.

As a base model, we have used 50 layers deep convolution neural network, which is ResNet50. ResNet50 was trained on more than a million images in the ImageNet dataset. The pre-trained network can classify images into 1000 categories, but we need 32 object categories such as Fried Rice, Chotpoti, Subway, etc. Pre-trained ResNet50 does not classify images by those 32 specific categories. We created a new model from scratch for this specific purpose, but for good results, we would need many images with labels for determining Fried Rice and Chotpoti, etc.

flatten layer and a new prediction layer for our required 32 classes.

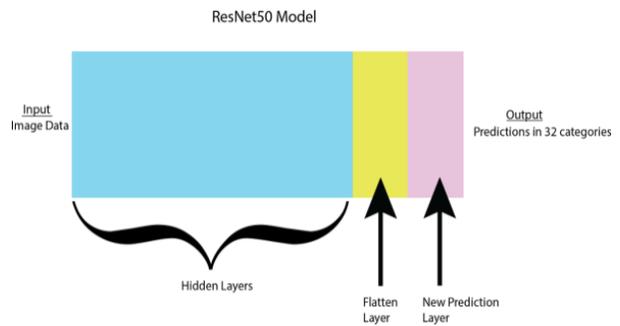

Fig. 2. Block diagram for proposed Transfer Learning model

Some layers before that in the pre-trained model may identify features like roads, buildings, windows, etc. will drop in a replacement for the last layer of ResNet50 model, this new last layer will predict whether an image is Chotpoti or Subway or any other from 30 classes.

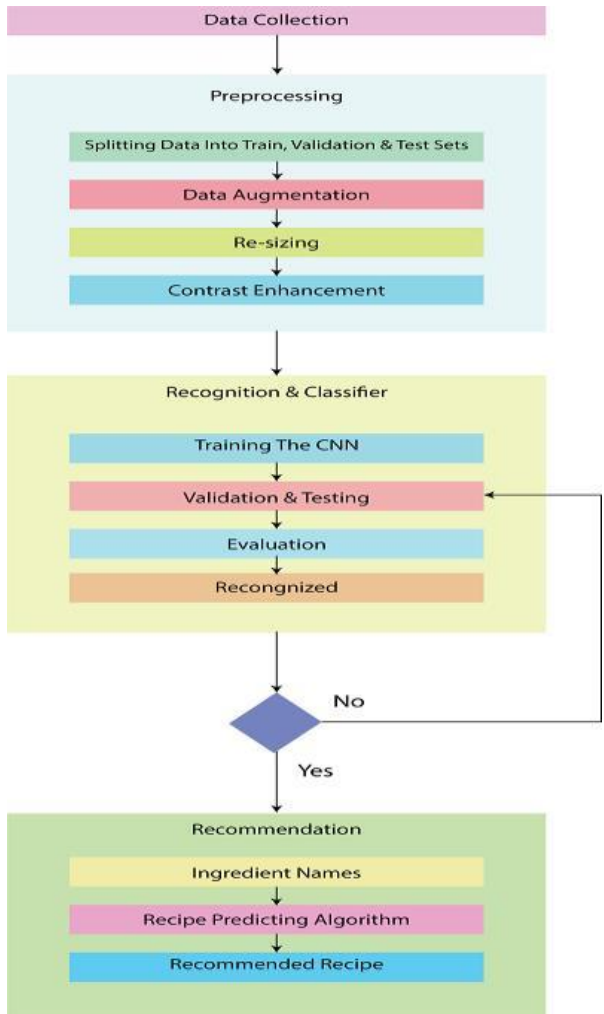

Fig. 1. Block diagram of the proposed method

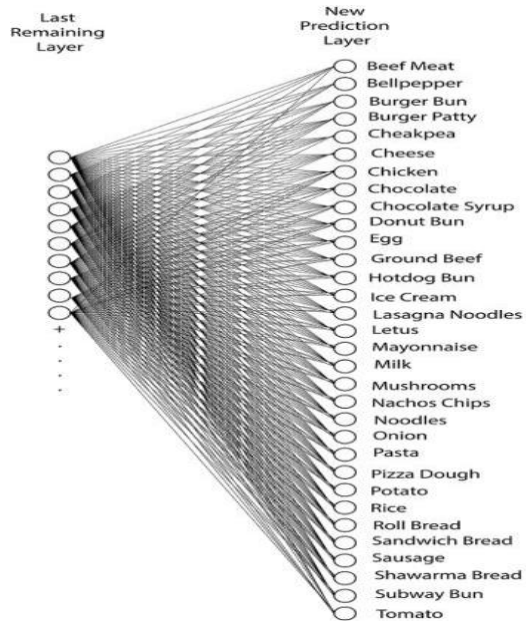

Fig. 3. Connection between last remaining layer and new prediction layer

It was possible to gain good results with fewer data using transfer learning. That's why we chose transfer learning. An early layer of a deep learning model identifies shapes, a last layer identifies more complex visual patterns, and the final layer makes predictions. Due to the similar low-level patterns involved in most computer vision problems, most layer layers of a pre-trained model are helpful for new applications. This means that most of the layers of the pre-trained ResNet50 models can be reused and we need to replace only that final layer that is used to generate predictions. That is why we've cut down the last prediction layer of the ResNet50 model trained for 1000 classes and added a

We have a lot of connections here. In figure 3, we can see the ResNet50 model has many layers, we have just cut off the last layer. What's left in the last layer is information about our image content stored as a series of numbers in a tensor. It should be a one-dimensional tensor, also known as a vector. It can be shown as a series of dots. Dots are called nodes. The first node represents the first number in the vector and the second node represents the second number and so on. We want to classify the images into 32 categories: Beef Meat, Bell pepper, Burger Bun, Burger Patty, Chickpea, Cheese, Chicken, Chocolate, Chocolate Syrup, Donut Bun, Egg, Ground Beef, Hotdog Bun, Ice Cream, Lasagna Noodles, Letus, Mayonnaise, Milk, Mushrooms, Nachos Chips, Noodles, Onion, Pasta, Pizza Dough, Potato, Rice, Roll Bread, Sandwich Bread, Sausage, Shawarma Bread, Subway Bun, and Tomato. In the last layer, we keep the pre-trained model and add another layer with 32 nodes. The first node capture how the Beef Meat image is, the second node capture how the Bell pepper image is, the third node capture how the Burger Bun image is, and so on. We've used training data to determine which nodes suggest an image is Burger Bun, which is Pasta, and so on.

## 3.2 Architecture

It is a variant of ResNet [7]. In ResNet 50 there are 48 convolution layers as well as 1 MaxPool and 1 Average Pool layer. In this network, there is a total of 50 layers

- The first layer will have a feature size of 7*7 and 64 such filters all with a stride of size 2. And it gives 1 layer.
- Next, we will see max pooling with the window size of 3*3 and a stride size of 2.
- 1 * 1, 64 means the first convolution layer has a filter size of 1 * 1, and 64 such filters. Next 3 * 3, 64 means filter size of 3 * 3, and 64 such filters and at last 1 * 1, 256 means, filter size of 1 * 1, and 256 such filters. These three layers are repeated 3 times which means 9 layers in this step.

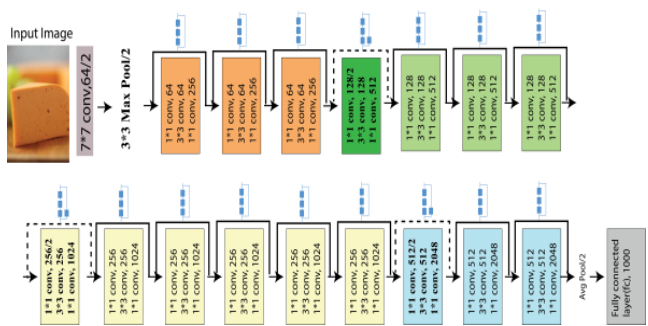

Fig. 4. Architecture of ResNet50

- After that 1*1,128 layer, then 3*3,128, and at the last of 1 * 1, 512 this step will be repeated 4 times so that gives us 12 layers in this step.
- Next 1*1, 256 layers, then 3*3, 256, and then 1*1, 1024, and this will be repeated 6 times so that gives us 18 layers.
- And then again a1*1, 512 layers, 3*3, 512 and 1* 1, 2048, and this was repeated 3 times so that it gives 9 layers.

Our next step is to use an average pool, and then finish it up with a fully connected layer of 1000 nodes. Lastly, there is a SoftMax function that provides us 1 layer. Max-pooling layers, along with activation functions, are not counted. In total, we have a 1+9+12+18+9+1=50 layers Deep Convolutional Network.

## 3.3 Recipe Recommendation

As mentioned before our aim is to recommend cooking recipes from recognized food ingredients. We have generated our recipe database and algorithm for recipe recommendations. We have selected 19 kinds of cooking recipes associated with 32 ingredients and designed a 2D matrix (19 rows × 32 columns) for the recipe recommendation algorithm in which 19 rows contain 19 recipes and 32 columns contain 32 food ingredients.

|  | Beef Meat | Bellpepper | Burger Bun | Burger Patty | Cheakpea | Cheese | Chicken | Chocolate | Chocolate syrup | Donut Bun | Egg | Ground Beef | Hotdog Bun | Ice Cream | Lasagna Noodles | Letus | Mayonnaise | Milk | Mushrooms | Nachos Chips | Noodles | Onion | Pizza Dough | Potato | Pasta | Rice | Roll Bread | Sandwich Bread | Sausage | Shawarma Bread | Subway Bun | Tomato |
|---|---|---|---|---|---|---|---|---|---|---|---|---|---|---|---|---|---|---|---|---|---|---|---|---|---|---|---|---|---|---|---|---|
| Burger | 0 | 0 | 1 | 1 | 0 | 1 | 0 | 0 | 0 | 0 | 0 | 0 | 0 | 0 | 0 | 1 | 0 | 0 | 0 | 0 | 0 | 0 | 0 | 0 | 0 | 0 | 0 | 0 | 0 | 0 | 0 | 0 |
| Chicken Fry | 0 | 0 | 0 | 0 | 0 | 0 | 1 | 0 | 0 | 0 | 0 | 0 | 0 | 0 | 0 | 0 | 0 | 0 | 0 | 0 | 0 | 0 | 0 | 0 | 0 | 0 | 0 | 0 | 0 | 0 | 0 | 0 |
| Chicken Wings | 0 | 0 | 0 | 0 | 0 | 0 | 1 | 0 | 0 | 0 | 0 | 0 | 0 | 0 | 0 | 0 | 0 | 0 | 0 | 0 | 0 | 0 | 0 | 0 | 0 | 0 | 0 | 0 | 0 | 0 | 0 | 0 |
| Chocolate Milkshake | 0 | 0 | 0 | 0 | 0 | 0 | 0 | 1 | 1 | 0 | 0 | 0 | 0 | 1 | 0 | 0 | 0 | 1 | 0 | 0 | 0 | 0 | 0 | 0 | 0 | 0 | 0 | 0 | 0 | 0 | 0 | 0 |
| Chotpoti | 0 | 0 | 0 | 0 | 1 | 0 | 0 | 0 | 0 | 0 | 1 | 0 | 0 | 0 | 0 | 0 | 0 | 0 | 0 | 0 | 0 | 0 | 0 | 1 | 0 | 0 | 0 | 0 | 0 | 0 | 0 | 0 |
| Donut | 0 | 0 | 0 | 0 | 0 | 0 | 0 | 1 | 0 | 1 | 0 | 0 | 0 | 0 | 0 | 0 | 0 | 0 | 0 | 0 | 0 | 0 | 0 | 0 | 0 | 0 | 0 | 0 | 0 | 0 | 0 | 0 |
| French Fries | 0 | 0 | 0 | 0 | 0 | 0 | 0 | 0 | 0 | 0 | 0 | 0 | 0 | 0 | 0 | 0 | 0 | 0 | 0 | 0 | 0 | 0 | 0 | 1 | 0 | 0 | 0 | 0 | 0 | 0 | 0 | 0 |
| Fried Rice | 0 | 0 | 0 | 0 | 0 | 0 | 1 | 0 | 0 | 0 | 1 | 0 | 0 | 0 | 0 | 0 | 0 | 0 | 0 | 0 | 0 | 0 | 0 | 0 | 0 | 1 | 0 | 0 | 0 | 0 | 0 | 0 |
| Hot Dogs | 0 | 0 | 0 | 0 | 0 | 0 | 0 | 0 | 0 | 0 | 0 | 0 | 1 | 0 | 0 | 0 | 0 | 0 | 0 | 0 | 0 | 0 | 0 | 0 | 0 | 0 | 0 | 0 | 1 | 0 | 0 | 0 |
| Lasagna | 0 | 1 | 0 | 0 | 0 | 1 | 0 | 0 | 0 | 0 | 0 | 1 | 0 | 0 | 1 | 0 | 0 | 0 | 1 | 0 | 0 | 1 | 0 | 0 | 0 | 0 | 0 | 0 | 0 | 0 | 0 | 1 |
| Nachos | 1 | 0 | 0 | 0 | 0 | 1 | 0 | 0 | 0 | 0 | 0 | 0 | 0 | 0 | 0 | 0 | 0 | 0 | 0 | 1 | 0 | 0 | 0 | 0 | 0 | 0 | 0 | 0 | 0 | 0 | 0 | 0 |
| Noodles | 0 | 0 | 0 | 0 | 0 | 0 | 1 | 0 | 0 | 0 | 1 | 0 | 0 | 0 | 0 | 0 | 0 | 0 | 0 | 0 | 1 | 0 | 0 | 0 | 0 | 0 | 0 | 0 | 0 | 0 | 0 | 0 |
| Pasta | 0 | 0 | 0 | 0 | 0 | 1 | 1 | 0 | 0 | 0 | 0 | 0 | 0 | 0 | 0 | 0 | 0 | 0 | 0 | 0 | 0 | 0 | 0 | 0 | 1 | 0 | 0 | 0 | 0 | 0 | 0 | 0 |
| Pizza | 1 | 1 | 0 | 0 | 0 | 1 | 0 | 0 | 0 | 0 | 0 | 0 | 0 | 0 | 0 | 0 | 0 | 0 | 1 | 0 | 0 | 1 | 1 | 0 | 0 | 0 | 0 | 0 | 1 | 0 | 0 | 0 |
| Potato Wedges | 0 | 0 | 0 | 0 | 0 | 0 | 0 | 0 | 0 | 0 | 0 | 0 | 0 | 0 | 0 | 0 | 0 | 0 | 0 | 0 | 0 | 0 | 0 | 1 | 0 | 0 | 0 | 0 | 0 | 0 | 0 | 0 |
| Roll | 0 | 0 | 0 | 0 | 0 | 0 | 1 | 0 | 0 | 0 | 0 | 0 | 0 | 0 | 0 | 0 | 0 | 0 | 0 | 0 | 0 | 0 | 0 | 0 | 0 | 0 | 1 | 0 | 0 | 0 | 0 | 0 |
| Sandwich | 0 | 0 | 0 | 0 | 0 | 1 | 1 | 0 | 0 | 0 | 0 | 0 | 0 | 0 | 0 | 1 | 0 | 0 | 0 | 0 | 0 | 0 | 0 | 0 | 0 | 0 | 0 | 1 | 0 | 0 | 0 | 0 |
| Shawrma | 0 | 0 | 0 | 0 | 0 | 0 | 1 | 0 | 0 | 0 | 0 | 0 | 0 | 0 | 0 | 0 | 0 | 0 | 0 | 0 | 0 | 0 | 0 | 0 | 0 | 0 | 0 | 0 | 0 | 1 | 0 | 0 |
| Subway | 0 | 0 | 0 | 0 | 0 | 0 | 1 | 0 | 0 | 0 | 0 | 0 | 0 | 0 | 0 | 0 | 0 | 0 | 0 | 0 | 0 | 0 | 0 | 0 | 0 | 0 | 0 | 0 | 0 | 0 | 1 | 0 |

Fig. 5. 2D matrix (19 rows × 32 columns) for recipe recommendation algorithm.

In figure 5, we created a one-to-many relationship with food ingredients and recipes. We have objectified here which recipes can be cooked with which ingredients. For example, Egg is needed for cooking Chotpoti, Fried Rice, and Noodles. So we have assigned "1" in those rows where recipes can be cooked with Egg and the rest of the rows remain "0" for Egg. It is like binary classification. Chicken is needed for the highest number of recipes. Nine recipes can be cooked with Chicken, those are Chicken Fry, Chicken Wings, Fried Rice, Noodles, Pasta, Roll, Sandwich, Shawarma, and Subway. So we have assigned "1" in these recipe rows against Chicken and the other ten rows for recipes remain "0" against Chicken.

After food ingredients recognition using our developed CNN model, we've run a linear search in this database (19 × 32 matrices). For example, our developed CNN model recognized three food ingredients which are Chicken, Egg, and Rice. So our developed algorithm will conduct a linear search in this database and cross-check which recipes are assigned "1" for Chicken, Egg, and Rice food ingredients columns and other food ingredients columns are assigned "0". Then it will find Fried Rice. Multiple recipes can be recommended by our developed algorithm. For example, the recognized food ingredients are Chickpea, Egg, and Potato. First of all the recipe recommendation algorithm will conduct a linear search in the database for Chickpea, Egg, and Potato individually and cross- check which recipes are assigned "1" only for Chickpea or Egg or Potato and then cross-check which recipes are assigned "1" for Chickpea, Egg and Potato altogether. Then it will find Hotpot, French Fries, and Potato Wedges. This is how our developed recipe recommendation algorithm works.

### 3.3 Dataset

We had to create our own since there were no real-life datasets available online for this topic, such as pizza dough, lettuce, mayonnaise, etc. For this study, we have used three datasets from Kaggle like food101, fruit 360, and UECFOOD256 and the data we created. From fruit360, we took the potato and tomato datasets, and from food101, we took the ice cream and onion datasets, and finally, from UECFOOD256, we took the rice, chicken, roll bread, and sausage datasets. In total, we made the dataset which contains a total of 9856 images of 32 different ingredients in figure 6.

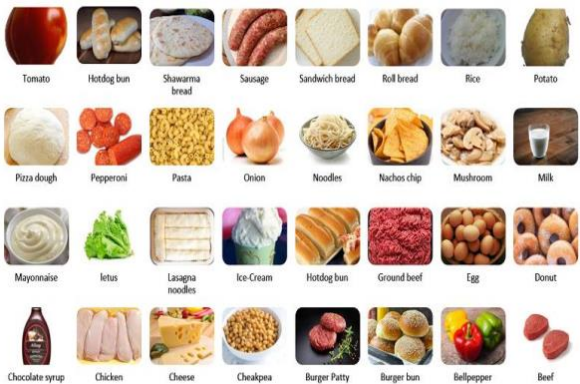

Fig. 6. 32 kinds of food ingredients in our dataset

### 3.4 Preprocessing

Since convolution neural networks accept inputs of the same size, all the images of food ingredients should be resized to a discerned size before inputting them to the CNN. The targeted input size for ResNet50 was 224 × 224. But our dataset images were in different shapes and sizes as we gathered them from different sources. Some food ingredients datasets were available free on Kaggle and other datasets we created our own by a camera. Training a CNN on unreformed images will surely lead to terrific classification exhibitions. So we have resized the images into 224 × 224 pixels and excluded unnecessary objects from the images.

### 3.5 Data Augmentation

A machine learning model performs better and is more accurate when the dataset is affluent and adequate. We've got pre-made datasets of onion, tomato, potato, and rice among all the ingredients. We've used data augmentation to increase the number of photos for Beef Meat, Bell Pepper, Burger Bun, Burger Patty, Chickpea, Cheese, Chicken, Chocolate Syrup, Donut Bun, Ground Beef, Hotdog Bun, Lasagna Noodles, Lettuce, Mayonnaise, Milk, Mushrooms, Nachos Chips, Noodles, Pasta, Egg, Pizza Dough, Chocolate, Roll Bread, Sandwich Bread, Sausage, Shawarma Bread, and Subway Bun. We used the TensorFlow library to perform data augmentation. The Image Data Generator function was used to supplement data that is a Keras function. During data augmentation, images have been rotated 45 degrees, shifted width and height by 20%, zoomed by 20%, flipped horizontally, and sheared with a range of 20%. Image data augmentation is usually applied only to the training dataset, not the validation or assessment datasets.

### 3.6 Optimizer and Learning Rate

In optimization, optimizers are algorithms or methods that modify deep neural network attributes, such as weights and learning rates, to reduce losses. Optimizers are used to minimize a function as part of solving optimization problems. There are various kinds of optimizers available as like RMSprop, Adam, Nadam, SGD, Ftrl etc. Adam optimizer [21] is used in our research. Adam optimization is a technique that extends stochastic gradient descent for use in computer vision and natural language processing. It is commonly used by researchers in PC vision because of its better performance. In our model Adam optimizer is used with the learning rate of 0.001. It's the default learning rate of Adam optimizer. In training a convolution neural network, the learning rate plays a significant role. A lower learning rate will yield better classification; however, the optimizer will set aside more effort to achieve global optima, which will reduce the loss. The higher the learning rate, the worse the accuracy. 0.001 was set as the learning rate as that would naturally drop upon checking the validation accuracy. For calculating the loss we used the Categorical cross- entropy function. Categorical cross-entropy is also known as Softmax loss. This is a Softmax activation with a cross- entropy loss as well. In comparison to other functions, like classification errors, mean squared errors, etc., we found that cross-entropy shows a good quality [22] since our model is a multiclass classification problem, Categorical cross-entropy is the fittest choice for us.

## 4 RESULTS AND DISCUSSION

### 4.1 Model Performance

As we mentioned, we have a dataset of a total of 9856 images. So, we took 70% of the images for the training, 20% for testing, and the rest of the 10% for validation. After running 20 epochs using transfer learning, our proposed CNN-based ResNet50 model achieved a 99.71% accuracy for the training dataset. And 92.6% accuracy on the validation dataset. After analyzing the results and confusion matrix, it can be stated that the CNN model has acceptable performance for classifying food ingredients. The CNN model's performance is shown below in figure 7 and 8.

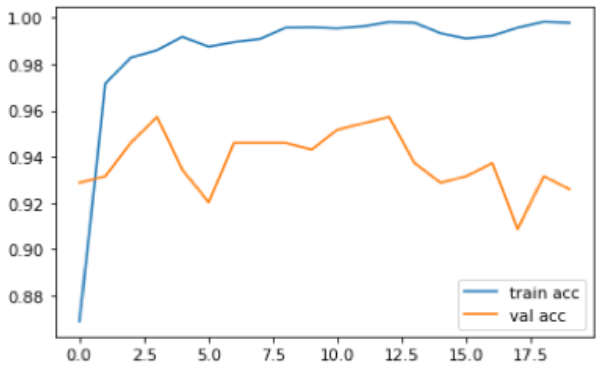

Fig. 7. Training and Validation accuracy.

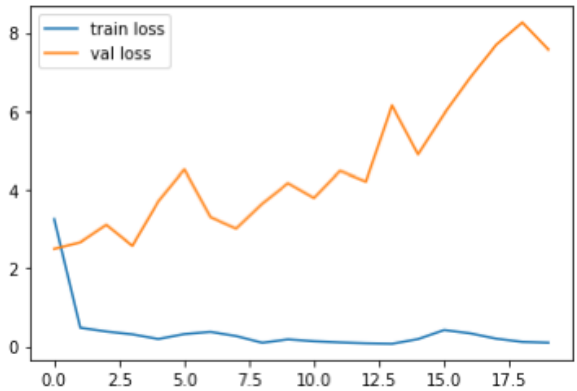

Fig. 8. Training and Validation loss

### 4.2 Result Comparison

Research for recipe recommendations can hardly be seen. There are some papers on food item recognition that are roughly like our study.

TABLE 1. COMPARISON BETWEEN SOME PREVIOUS WORKS

| Work | Technology used | Accuracy |
| --- | --- | --- |
| Khan, et al. "A machine learning approach to recognize junk food (proposed Model)" [23] | CNN | 90.47% |
| Sun, et al. "Yelp Food Identification via Image Feature Extraction and Classification" [24] | CNN &SVM | 68.49% |
| Yanai, et al. "Food image recognition using deep convolutional network with pre- training and fine tuning" [25] | DCNN | 70.4% |
| Baxter, et al. "Food recognition using ingredient-level features" [26] | CNN | 81.2% |
| R. Xu, et al "Modeling restaurant context for food recognition" [27] | CNN | 84.9% |
| Ciocca, et al. "Food recognition: a new dataset, experiments, and results" [28] | CNN | 79% |
| Mayers, et al. "Im2calories: towards an automated mobile vision food diary" [29] | Googlre Net | 79% |
| **Recipe recommendation based on ingredients detection using deep learning** | **CNN-based Resnet50** | **94%** |

## 4 CONCLUSION

This paper presented a CNN model to recognize food ingredients and a recipe recommendation algorithm based on detected ingredients to suggest cooking recipes. In addition, we introduced a custom dataset of 32 categories of food ingredients. We achieved the testing accuracy of 94%, which is impressive, and proved that the performance of this model to recognize food ingredients from images is more advanced. Due to these high accuracy levels, we found CNN to be particularly suitable for food ingredients recognition. Moreover, if we could get a high configuration of computer GPU, then the accuracy level could be above the current accuracy level as mentioned before; we used transfer learning for training the CNN model. ResNet50 was used as a base model. As part of an experiment, we also used VGG16 and MobileNetV2 as a base model, but ResNet50 generated better performance than the other two pre-trained models. We selected 19 cooking recipes against 32 food ingredients and developed a unique algorithm for recipe recommendations based on ingredients detection.

In the future, we aim to increase the number of recipes and food ingredients and enrich our dataset. We've found two problems in our dataset; one is contrast variation of images, and another is imbalance problem of classes. We used data augmentation to reduce the imbalance problem. In the future, we will work to overcome these two problems completely. We have got a limitation in our research: our CNN model cannot classify multiple objects. That means our CNN model can recognize a single food ingredient at a time. It cannot identify various food ingredients from a given image. We want to deploy multiple object detection in our research in the future.